%% file: paper1341.tex
\documentclass[runningheads]{llncs}
\usepackage{subcaption}
\usepackage{tabularx}

\usepackage{amssymb, amsmath, amsfonts}
\usepackage{acro}
\input{acronyms}

\usepackage{graphicx}
\usepackage{hyperref, xcolor}

\usepackage[misc,geometry]{ifsym} 

\begin{document}
\title{Multimodal brain age estimation using interpretable adaptive population-graph learning}
\titlerunning{Adaptive population-graph learning for brain age estimation}

\author{Kyriaki-Margarita Bintsi \inst{1} \textsuperscript{(\Letter)}, Vasileios Baltatzis \inst{1, 2}, Rolandos Alexandros Potamias \inst{1}, Alexander Hammers \inst{2}, Daniel Rueckert \inst{1,3}}
% index {Bintsi, Kyriaki-Margarita}
% index {Baltatzis, Vasileios}
% index {Potamias, Rolandos Alexandros}
% index {Hammers, Alexander}
% index {Rueckert, Daniel}
\authorrunning{K. M. Bintsi et al.}

\institute{Department of Computing, Imperial College London, UK \\
\email{m.bintsi19@imperial.ac.uk}\\ 
\and Biomedical Engineering and Imaging Sciences, King's College London, UK \\
\and Technical University of Munich, Germany
}

\maketitle              % typeset the header of the contribution
\begin{abstract}
% The abstract should briefly summarize the contents of the paper in
% 15--250 words.
Brain age estimation is clinically important as it can provide valuable information in the context of neurodegenerative diseases such as Alzheimer's. Population graphs, which include multimodal imaging information of the subjects along with the relationships among the population, have been used in literature along with \acp{gcn} and have proved beneficial for a variety of medical imaging tasks. A population graph is usually static and constructed manually using non-imaging information. However, graph construction is not a trivial task and might significantly affect the performance of the \ac{gcn}, which is inherently very sensitive to the graph structure. In this work, we propose a framework that learns a population graph structure optimized for the downstream task. An attention mechanism assigns weights to a set of imaging and non-imaging features (phenotypes), which are then used for edge extraction. The resulting graph is used to train the \ac{gcn}. The entire pipeline can be trained end-to-end. Additionally, by visualizing the attention weights that were the most important for the graph construction, we increase the interpretability of the graph. We use the UK Biobank, which provides a large variety of neuroimaging and non-imaging phenotypes, to evaluate our method on brain age regression and classification. The proposed method outperforms competing static graph approaches and other state-of-the-art adaptive methods. We further show that the assigned attention scores indicate that there are both imaging and non-imaging phenotypes that are informative for brain age estimation and are in agreement with the relevant literature.

\end{abstract}

\section{Introduction}
% Aging - pad
Healthy brain aging follows specific patterns \cite{alam2014morphological}. However, various neurological diseases, such as Alzheimer's disease \cite{davatzikos2011prediction}, Parkinson's disease \cite{reeve2014ageing}, and schizophrenia \cite{koutsouleris2014accelerated}, are accompanied by an abnormal accelerated aging of the human brain. Thus, the difference between the biological brain age of a person and their chronological age can show the deviation from the healthy aging trajectory, and may prove to be an important biomarker for neurodegenerative diseases \cite{cole2017predicting,franke2019ten}. 

Recently, graph-based methods have been explored for brain age estimation as graphs can inherently combine multimodal information by integrating the subjects' neuroimaging information as node features and, through a similarity metric, the associations among subjects through as edges that connect these nodes \cite{parisot2018disease}. However, in medical applications, the construction of a population-graph is not always simple as there are various ways subjects could be considered similar.

% GNNs for classification tasks - Static graphs, parisot etc.
\acp{gcn} \cite{kipf2016semi} have been extensively \cite{ahmedt2021graph} used in the medical domain for node classification tasks, such as Alzheimer's prediction \cite{parisot2018disease,kazi2019inceptiongcn} and Autism classification \cite{anirudh2019bootstrapping}. They take graphs as input and, in most cases, the graph structure is predefined and static. 
\ac{gcn} performance is highly dependent on the graph structure. 
This has been correlated in related literature with the heterophily of graphs in the semi-supervised node classification tasks, which refers to the case when the nodes of a graph are connected to nodes that have dissimilar features and different class labels \cite{zheng2022graph}. It has been shown that if the homophily ratio is very low, a simple \ac{mlp} that completely ignores the structure, can outperform a \ac{gcn} \cite{zhu2020beyond}.

\begin{figure}[t]
\includegraphics[width=\textwidth]{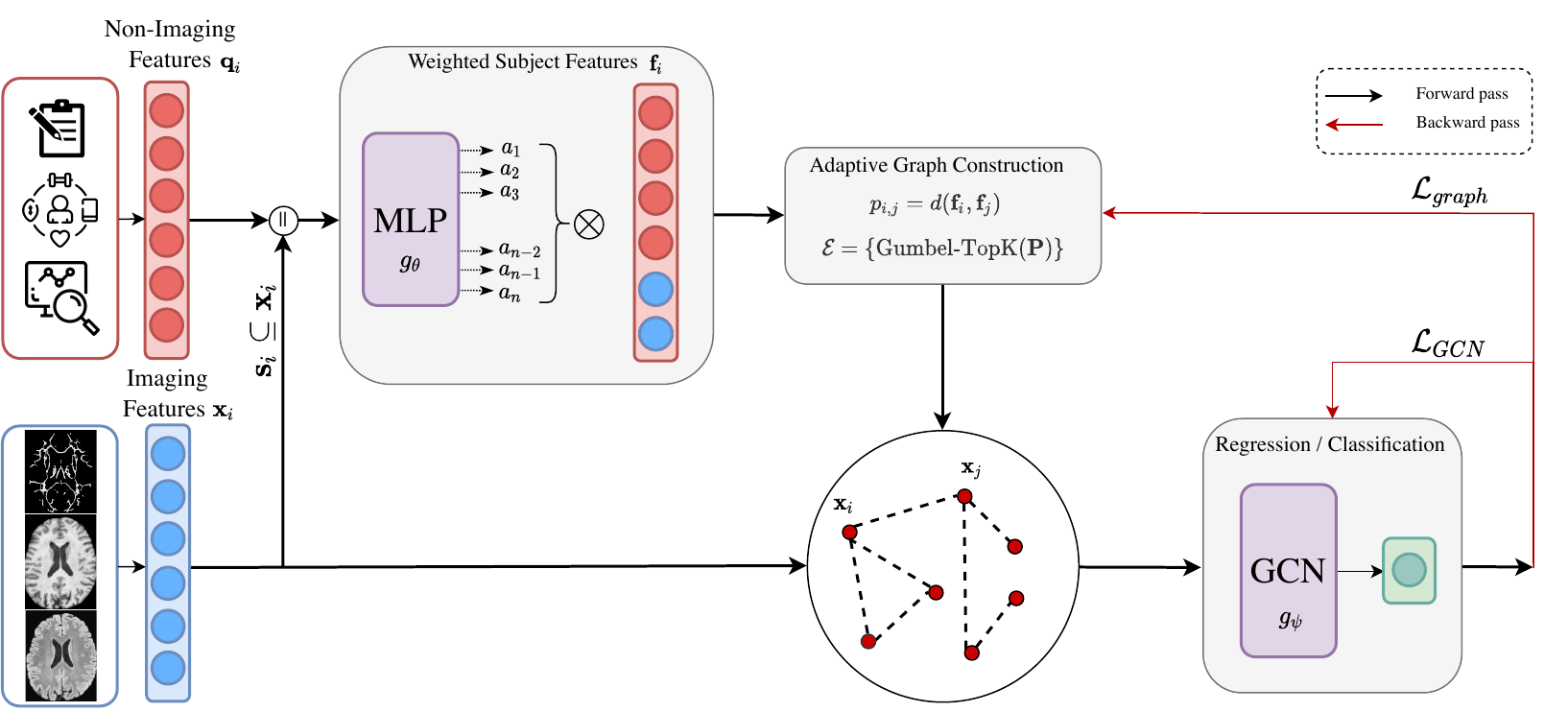}
\caption{Proposed methodology: The non-imaging features $\textbf{q}_i$ and a subset of the imaging features $\textbf{s}_i \subseteq \textbf{x}_i$ per subject $i$ are used as input to a \ac{mlp}, which produces attention weights for each one of these features. Based on these, the edges of the graph are stochastically sampled using the Gumbel-Top-k trick. The constructed graph, which uses the imaging features $\textbf{X}$ as node features is used to train a \ac{gcn} for brain age estimation tasks.} \label{fig:pipeline}
\end{figure}

% GNNs for graph learning - kazi, cosmo, ia-gcn, mmgl, ev-gcn.
% Not using phenotypes - not understandable graph.
A way to address this problem is through adaptive graph learning \cite{wei2022graph}, which learns the graph structure through training. In the medical domain, there is little ongoing research on the topic \cite{kazi2022differentiable,zheng2022multi}.
However, the adaptive graphs in \cite{kazi2022differentiable,kazi2021ia,cosmo2020latent} are connected based on the imaging features and do not take advantage of the associations of the non-imaging information for the edges. In \cite{huang2020edge}, non-imaging information is used for graph connectivity, but the edges are already pre-pruned similar to \cite{parisot2018disease}, and only the weights of the edges can be modified. In \cite{wang2019dynamic}, even though the graph is dynamic, it is not being learnt.
% Maybe talk about interpretability on graphs??
While the extracted graphs in these studies are optimized for the task at hand, there is no explanation for the node connections that were proposed. Given that interpretability is essential in the medical domain \cite{shaban2021explainability}, since the tools need to be trusted by the clinicians, an adaptive graph learnt during the training, whose connections are also interpretable and clear can prove important.
 % GNNs for age estimation (kamilest?) 
Additionally, most existing works focus on brain age classification in four bins, usually classifying the subjects per decade \cite{kazi2022differentiable,kazi2021ia,cosmo2020latent}. Age regression, which is a more challenging task, has not been extensively explored with published results not reaching sufficient levels of performance \cite{stankeviciute2020population}. 
% However, the performance is highly depended on the neuroimaging features used so these studies are not directly comparable. 

\noindent{\textbf{Contributions.}} This paper has the following contributions: 1) We combine imaging and non-imaging information in an attention-based framework to learn adaptively an optimized graph structure for brain age estimation. 2) We propose a novel graph loss that enables end-to-end training for the task of brain age regression. 3) Our framework is inherently interpretable as the attention mechanism allows us to rank all imaging and non-imaging phenotypes according to their significance for the task. 4) We evaluate our method on the \ac{ukbb} and achieve state-of-the-art results on the tasks of brain age regression and classification. The code can be found on GitHub at: 
\url{https://github.com/bintsi/adaptive-graph-learning}.
% The code will be made publicly available.
% We do so by giving trainable attention weights to the phenotypes, which we then use to extract the edges that are used as input to a \ac{gcn}. We implement the proposed approach both for age classification and age regression. The network is trained end-to-end, as we believe that the use of the attention weights will not only improve the performance, but also give an insight about the phenotype that are most relevant to the task of age estimation, thus making the pipeline more interpretable. We also include several ablation studies to demonstrate how the choice of the phenotypes used, and the different ways to connect the edges, may affect the performance of the \ac{gcn}. 

\section{Methods}

% Mathematical definition of graphs, node features, edges/adjacency matrix, labels, phenotypes.
Given a set of $N$ subjects with $M$ features $\textbf{X} = [ \mathbf{x}_1, ..., \mathbf{x}_N] \in \mathbb{R}^{N \times M}$ and labels $\textbf{y} \in \mathbb{R}^N$, a population graph is defined as $\mathcal{G}=\{\mathcal{V}, \mathcal{E}\}$, where $\mathcal{V}$ is a set of nodes, one per subject, and $\mathcal{E}$ is a set of paired nodes that specifies the connectivity of the graph, meaning the edges of the graph.
To create an optimized set of edges for the task of brain age estimation, we leverage a set of non-imaging phenotypes $\textbf{q}_i \in \mathbb{R}^Q$ and a set of imaging phenotypes $\textbf{s}_i \in \mathbb{R}^S$ per subject $i$ through an attention-based framework. The imaging phenotypes are a subset of the imaging features $\textbf{s}_i \subseteq \textbf{x}_i$. The phenotypes are selected according to \cite{cole2020multimodality}. The resulting graph is used as input in a \ac{gcn} that performs node-level prediction tasks. Here, we give a detailed description of the proposed architecture, in which the connectivity of the graph $\mathcal{E}$, is learnt through end-to-end training in order to find the optimal graph structure for the task of brain age estimation. An outline of the proposed pipeline may be found in Figure \ref{fig:pipeline}.

\noindent{\textbf{Attention Weights Extraction.}}
% Phenotypes and subset of the imaging features to an MLP. Weights are extracted per phenotype. Differenriable. Outputs from 0 to 1.
Based on the assumption that not all phenotypes are equally important for the construction of the graph, we train a \ac{mlp} $g_\theta$, with parameters $\theta$, which takes as input both the non-imaging features $\textbf{q}_i$  and the imaging features $\textbf{s}_i$ for every subject $i$ and outputs an attention weight vector $\textbf{a} \in \mathbb{R}^{Q+S}$, where every element of \textbf{a} corresponds to a specific phenotype. Intuitively, we expect that the features that are relevant to brain age estimation will get attention weights close to 1, and close to 0 otherwise.
Since we are interested in the overall relevance of the phenotypes for the task, the output weights need to be global and apply to all of the nodes. To do so, we average the attention weights across subjects and normalize them between 0 and 1.

\noindent{\textbf{Edge Extraction.}}
The weighted phenotypes for each subject $i$ are calculated as $\textbf{f}_i=\textbf{a} \odot (\textbf{q}_i \mathbin\Vert \textbf{s}_i)$, where $\textbf{f}_i \in \mathbb{R}^{Q+S}$, $\textbf{a}$ are the attention weights produced by the \ac{mlp} $g_\theta$, $(\cdot\mathbin\Vert\cdot)$ denotes the concatenation function between two vectors and $\odot$ denotes the Hadamard product. We further define the probability $p_{ij}(\textbf{f}_i, \textbf{f}_j;\theta,t)$ of a pair of nodes $(i,j) \in \mathcal{V}$ to be connected in Equation \eqref{eq:prob}: 

\begin{equation}
    p_{ij}(\textbf{f}_i, \textbf{f}_j; \theta, t) =e^{-td{(\textbf{f}_i, \textbf{f}_j)}^2}
    \label{eq:prob}
\end{equation}

where $t$ is a learnable parameter and $d$ is a distance function that calculates the distance between the weighted phenotypes of two nodes. 
To keep the memory cost low, we create a sparse k-degree graph. We use the Gumbel-Top-k trick \cite{kool2019stochastic}, which acts as a stochastic relaxation of the kNN rule, in order to sample k edges for every node according to the probability matrix $\textbf{P} \in \mathbb{R}^{N \times N}$. Since there is stochasticity in the sampling scheme, multiple runs are performed at inference time and the predictions are averaged.

\noindent{\textbf{Optimization.}}
% Node features along with adjacency input to a GNN. 
The extracted graph is used as input, along with the imaging features $\textbf{X}$, which are used as node features, to a \ac{gcn} $g_{\psi}$, with parameters $\psi$, which comprises of a number of graph convolutional layers, followed by fully connected layers. 
% Output -> age estimation, can be both classification or regression
% The output of the final layer can be either 1, in the case of regression, or the number of classes, in the case of classification. 
% Loss function
The pipeline is trained end-to-end with a loss function $\mathcal{L}$ that consists of two components and is defined as 
in Equation \eqref{eq:loss}.
\begin{equation}
    \mathcal{L} = \mathcal{L}_{GCN} + \mathcal{L}_{graph}.
    \label{eq:loss}
\end{equation}
The first component, $\mathcal{L}_{GCN}$, is optimizing the \ac{gcn}, $g_{\psi}$. For regression we use the Huber loss \cite{huber1992robust}, while for classification we use the Cross Entropy loss function. The second component, $\mathcal{L}_{graph}$, optimizes the \ac{mlp} $g_{\theta}$, whose output are the phenotypes' attention weights. However, the graph is sparse with discrete edges and hence the network cannot be trained with backpropagation as is. To alleviate this issue, we formulate our graph loss in a way that rewards edges that lead to correct predictions and penalize edges that lead to wrong predictions. Inspired by \cite{kazi2022differentiable}, where a similar approach was used for classification, the proposed graph loss function is designed for regression instead and is defined in Equation \eqref{eq:graph_loss}:
% \begin{equation}
%     \mathcal{L}_{graph} = \sum_{\substack{i=1\\
%                                 j:(i,j) \in \mathcal{E}}}^N
%         \rho(y_i, f_{\psi}(X_i)) \, log(p_{ij}),
%     \label{eq:graph_loss}
% \end{equation}
\begin{equation}
    \mathcal{L}_{graph} = \sum_{
                                (i,j) \in \mathcal{E}}
        \rho(y_i, g_{\psi}(\textbf{x}_i)) \, log(p_{ij}),
    \label{eq:graph_loss}
\end{equation}
where $\rho(\cdot, \cdot)$ is the reward function which is defined in Equation \eqref{eq:reward}:
\begin{equation}
    \rho(y_i, g_{\psi}(\textbf{x}_i)) =  \lvert y_i - g_{\psi}(\textbf{x}_i) \rvert - \varepsilon 
    \label{eq:reward}
\end{equation}
Here $\varepsilon$ is the null model's prediction (i.e. the average brain age of the training set). Intuitively, when the prediction error $\lvert y_i - g_{\psi}(\textbf{x}_i) \rvert$ is smaller than the null model's prediction then the reward function will be negative. In turn, this will encourage the maximization of $p_{ij}$ so that $\mathcal{L}_{graph}$ is minimized.

\section{Experiments}

\begin{table}[t]
\caption{Performance of the proposed approach in comparison with a traditional machine learning model, static graph baselines, and the state-of-the-art for age regression (left), and 4-class age classification (right). (MAE in years)}
\parbox{.35\linewidth}{
\centering
\begin{tabular}{l||c|c}
\hline
Method &  MAE & $r$ score \\
\hline
Linear Regression & 3.82 & 0.75 \\
Static (Node features) & 3.87 & 0.75 \\
Static (Phenotypes)  & 3.98 & 0.74 \\
DGM & 3.72 & 0.75 \\
\textbf{Proposed}  & \textbf{3.61} & \textbf{0.79} \\
\hline
\end{tabular}
\label{tab1:regr}
}
\parbox{.75\linewidth}{
\centering
\begin{tabular}{l||c|c|c}
\hline
Method & Accuracy  & AUC & F1 \\
\hline
Logistic Regression & 0.54 & 0.79 & 0.54 \\
Static (Node features) & 0.53 & 0.75 & 0.52\\
Static (Phenotypes) & 0.52 & 0.75 & 0.51\\
DGM & 0.55 & \textbf{0.80} & 0.55\\
\textbf{Proposed} & \textbf{0.58} & \textbf{0.80} & \textbf{0.56}  \\
\hline
\end{tabular}
\label{tab1:class}
}
\hfill
\end{table}

\begin{table}[t]
\caption{Ablation studies. Left: Effect of the choice of phenotypes used for the connection of the edges. Right: Performance of the proposed using different distances on the phenotypes to connect the edges. (MAE in years)}
\parbox{.6\linewidth}{
\centering
\begin{tabular}{l||c|c}
\hline
No. of phenotypes & MAE & $r$ score \\
\hline
20 (non-imaging only) & 4.63 & 0.68  \\
\textbf{35 (non-imaging \& imaging)} & \textbf{3.61} &  \textbf{0.79} \\
50 (non-imaging \&  imaging) & 3.65 & 0.78 \\
68 (imaging only) & 3.73 & 0.77  \\
\hline
\end{tabular}
\label{tab1:phenotypes}
}
\parbox{.4\linewidth}{
\centering
\begin{tabular}{l||c|c}
\hline
Similarity  Metric &  MAE & $r$ score \\
\hline
Random & 5.59 & 0.61  \\
\textbf{Euclidean} & \textbf{3.61}  & \textbf{0.79}  \\
Cosine & 3.7  &  0.78 \\
Hyperbolic  & 4.08 & 0.72 \\
\hline
\end{tabular}
\label{tab1:distance}
}
\end{table}

\noindent{\textbf{Dataset.}}
The proposed framework is evaluated on the \ac{ukbb} \cite{Alfaro-Almagro2018}, which provides not only a wide collection of images of vital organs, including brain scans, but also various non-imaging information, such as demographics, and biomedical, lifestyle, and cognitive performance measurements. Hence, it is perfectly suitable for brain age estimation tasks that incorporate the integration of imaging and non-imaging information.
Here, we use 68 neuroimaging phenotypes and 20 non-imaging phenotypes proposed by \cite{cole2020multimodality} as the ones most relevant to brain age in \ac{ukbb}. The neuroimaging features are provided by \ac{ukbb}, and include measurements extracted from structural MRI and diffusion weighted MRI. All phenotypes are normalized from 0 to 1. 
The age range of the subjects is 47-81 years. We only keep the subjects that have available the necessary phenotypes ending up with about 6500 subjects. We split the dataset into 75\% for the training set, 5\% for the validation set, and 20\% for the test set. 
Our pipeline has been primarily designed to tackle the challenging regression task and therefore the main experiment is brain age regression. We also evaluate our framework on the 4-class classification task that has been used in other related papers. 

\noindent{\textbf{Baselines.}}
% Static Graphs Here 
% Describe static graphs and how we build the graph/create the adgjacency. KNN
% Using imaging info only
% Using phenotypes for the edges.
Given that a \ac{gcn} trained on a meaningless graph can perform even worse than a simple regressor/classifier, our first baseline is a Linear/Logistic Regression model. For the second baseline, we construct a static graph based on a similarity metric, more specifically cosine similarity, of a set of features (either node features or non-imaging and imaging phenotypes) using the kNN rule with $k=5$ and train a \ac{gcn} on that graph. Using euclidean distance as the similarity metric leads to worse performance and is therefore not explored for the baselines. We also compare our method with DGM, which is the state-of-the-art on graph learning for medical applications \cite{kazi2022differentiable}. Since this work is only applicable for classification tasks, we extend it to regression, implementing the graph loss function we used for our pipeline as well. 

\noindent{\textbf{Implementation Details.}}
The \ac{gcn} architecture uses ReLU activations and consists of one graph convolutional layer with 512 units and one fully connected layer with 128 units before the regression/classification layer.  The number and the dimensions of the layers are determined through hyperparameter search based on validation performance. The networks are trained with the AdamW optimizer \cite{loshchilov2017decoupled}, with a learning rate of 0.005 for 300 epochs and utilize an early stopping scheme. The average brain age of the training set, which is used in Equation \eqref{eq:graph_loss}, is $\varepsilon = 6$. We use PyTorch \cite{Paszke2019PyTorch:Library} and a Titan RTX GPU. The reported results are computed by averaging 10 runs with different initializations.

\begin{figure}[t]
\includegraphics[width=0.95\textwidth]{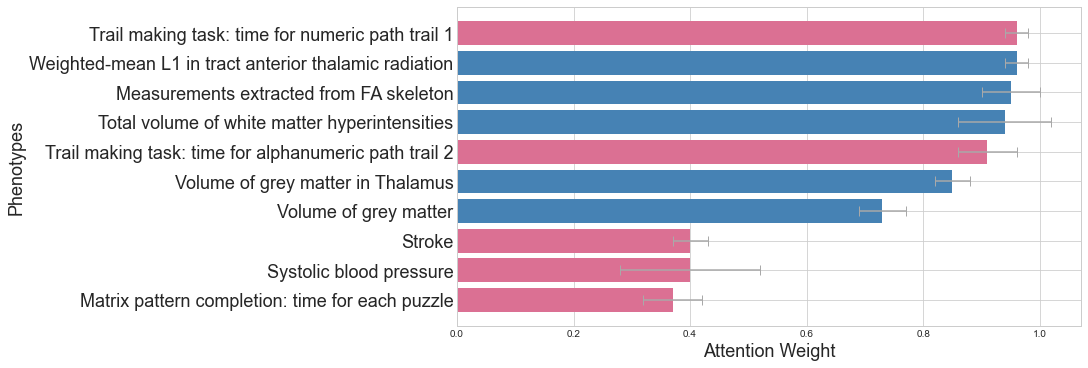}
\caption{Attention weights for the top 5 imaging (blue) and top 5 non-imaging phenotypes (pink) averaged across 10 runs of the pipeline.} \label{fig:top_scores}
\end{figure}

\subsection{Results}
\noindent{\textbf{Regression.}}
For the evaluation of the pipeline in the regression task, we use Mean Absolute Error (MAE), and the Pearson Correlation Coefficient ($r$ score).  A summary of the performance of the proposed and competing methods is available at Table \ref{tab1:regr}. Linear regression (MAE=3.82) outperforms GCNs trained on static graphs whether these are based on the node features (MAE=3.87) or leverage the phenotypes (MAE=3.98). The DGM outperforms the baselines (MAE=3.72). The proposed method outperforms all others (MAE=3.61).

\noindent{\textbf{Classification.}}
 For the classification task, we divide the subjects into four balanced classes. 
The metrics used for the classification task to evaluate the performance of the model are accuracy, the area under the ROC curve (AUC), which is defined as the average of the AUC curve of every class, and the Macro F1-score (Table \ref{tab1:class}).
A similar trend to the regression task appears here, with the proposed method reaching top performance with 58\% accuracy.

Our hypothesis that the construction of a pre-defined graph structure is suboptimal, and might even hurt performance, is confirmed. Both for the classification and regression tasks,  GCNs trained on static graphs do not outperform a simple linear model that ignores completely the structure of the graph. The proposed approach proves that not all phenotypes are equally important for the construction of the graph, and that giving attention weights accordingly increases the performance for both tasks. 

 \subsection{Ablation studies}
 
\noindent{\textbf{Number of phenotypes.}}
We perform an ablation test to investigate the effect of the number of features used for the extraction of the edges. We used only non-imaging phenotypes, only imaging phenotypes, or a combination of both (Table \ref{tab1:phenotypes}). The combination of imaging and non-imaging phenotypes (MAE=3.61) performs better than using either one of them, while adding more imaging phenotypes does not necessarily improve performance.
 
 \noindent{\textbf{Distance Metrics.}}
Moreover, we explore how different distance metrics affect performance. Here, euclidean, cosine, and hyperbolic \cite{krioukov2010hyperbolic}  distances are explored. We also include a random edge selection to examine whether using the phenotypic information improves the results. The results (Table \ref{tab1:distance}) indicate that euclidean distance performs the best, closely followed by cosine similarity. Hyperbolic performed a bit worse (MAE=4 years), while random edges gave a MAE of 5.59. Regardless of the distance metric,  phenotypes do incorporate valuable information as performance is considerably better than using random edges.

\begin{figure}[t]
\begin{subfigure}[t]{0.4\linewidth}
\centering
\includegraphics[scale=0.2]
{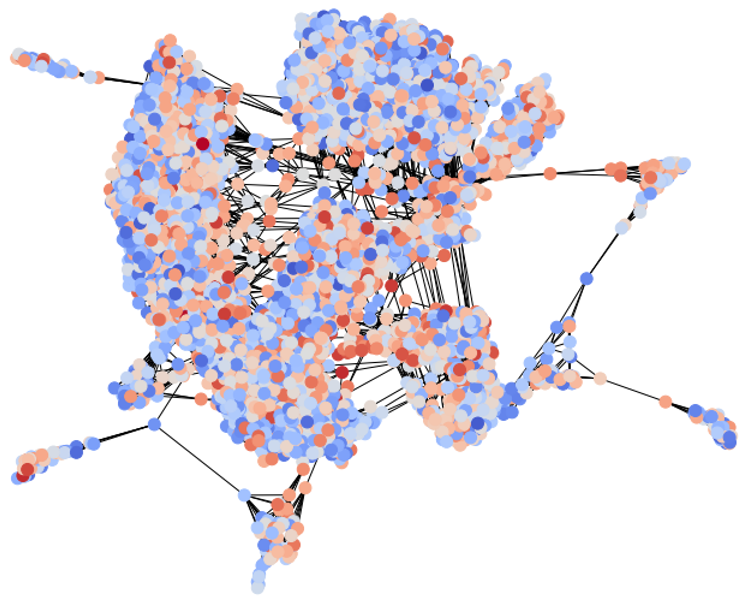}
\label{fig2a}
\end{subfigure}
\hfill
\begin{subfigure}[t]{0.4\linewidth}
\centering
\includegraphics[scale=0.2]
{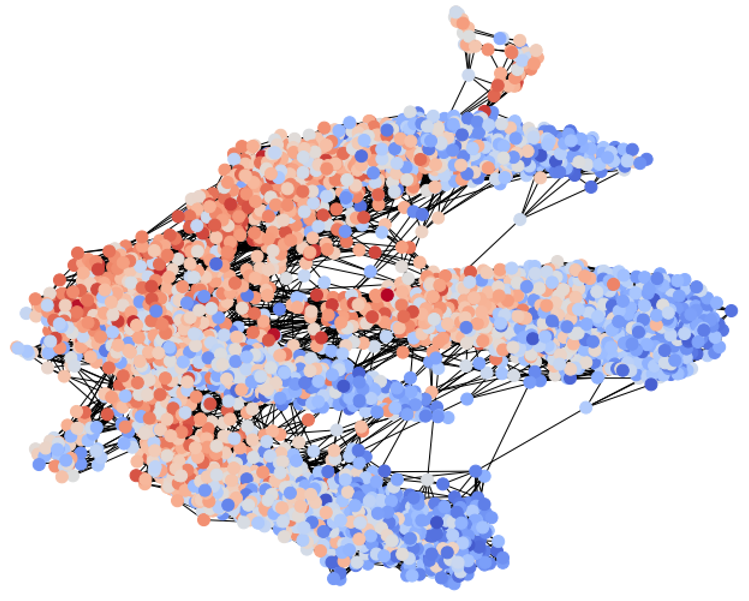}
\label{fig2b}
\end{subfigure}
\begin{subfigure}[t]{0.15\linewidth}
\centering
\includegraphics[scale=0.15]{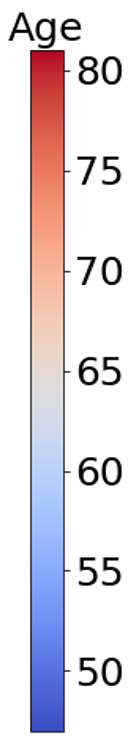}
\label{fig2c}
\end{subfigure}
\caption{Population-graph visualizations. Left: graph constructed based on the cosine similarity of the phenotypes. Right: graph constructed based on the cosine similarity of the weighted phenotypes, where the attention weights are the ones extracted from the proposed pipeline. The color of the nodes corresponds to the subject's age. It is evident that using the attention weights increases the homophily of the graph.} \label{fig:graphs}
\end{figure}

\subsection{Interpretability}
% We can find the important phenotypes and features.
% Visualization for the regression.
A very important advantage of the pipeline is that the graph extracted through the training is interpretable, in terms of why two nodes are connected or not. Visualizing the attention weights given to the phenotypes, we get an understanding of the features that are the most relevant to brain aging. 
The regression problem is clinically more important, thus we will focus on this in this section. A similar trend was presented for the classification task as well.

The imaging and non-imaging phenotypes that were given the highest attention scores in the construction of the graph can be seen in Figure \ref{fig:top_scores}. We color the non-imaging phenotypes in pink, and the imaging ones in blue. A detailed list of the names of the phenotypes along with the attention weights given by the pipeline can be found in the Supplementary Material. The non-imaging phenotypes that were given the highest attention weights, were two cognitive tasks, the numeric and the alphanumeric trail making tasks. Systolic blood pressure, and stroke, were the next most important non-imaging phenotypes, even though they were not as important as some of the imaging features. 
% The least relevant non-imaging phenotypes were the duration of the moderate activity, and the walking per week. 
Various neuroimaging phenotypes were considered important, such as information regarding the tract anterior thalamic radiaton, the volume of white matter hyperintensities, gray matter volumes, as well as measurements extracted from the FA (fractional anisotropy, a measure of the connectivity of the brain) skeleton. Our findings are in agreement with the relevant literature \cite{cole2020multimodality}. 

Apart from the attention weights, we also visualize the population graph that was used as the static graph of the baseline (Figure \ref{fig:graphs}(left)). This population graph is constructed based on the cosine similarity of the phenotypes, with all the phenotypes playing an equally important role for the connectivity. In addition, we visualize the population graph using the cosine similarity of the weighted phenotypes (Figure \ref{fig:graphs}(right)), with the attention weights provided by our trained pipeline. In both of the graph visualizations, the color of each node corresponds to the subject's age. It is clear that learning the graph through our pipeline results in a population graph where subjects with similar ages end up in more compact clusters, whereas the static graph does not demonstrate any form of organization. Since GCNs are affected by a graph construction with low homophily, it is reasonable that the static graphs perform worse than a simple machine learning method, and why the proposed approach manages to produce state-of-the-art results. Using a different GNN architecture that is not as dependent on the constructed graph's homophily could prove beneficial \cite{hamilton2017inductive,zheng2022graph}.

% Results table comments

% Comment attention weights

% The visualization of the weights shows that cognitive test measurements, volumes of gray matter, and white matter hyperintensities, and features extracted from the FA skeleton were found the most relevant to brain age

% Comment visualization

% % Ablation studies
% In the ablation studies, we experiment how the different spaces, euclidean, cosine, and hyperbolic, affect the performance.

% Limitations \& Future steps
\section{Conclusion}
In this paper, we propose an end-to-end pipeline for adaptive population-graph learning that is optimized for brain age estimation. We integrate multimodal information and use attention scores for the construction of the graph, while also increasing interpretability. The graph is sparse, which minimizes the computational costs. We implement the approach both for node regression and node classification and show that we outperform both the static graph-based and state-of-the-art approaches. We finally provide an insight into the most important phenotypes for graph construction, which are in agreement with related neurobiological literature. 

In future work, we plan to extract node features from the latent space of a CNN. Training end-to-end will focus on latent imaging features that are also important for the GCN. Such features would potentially be more expressive and improve the overall performance. Finally, we leverage the UKBB because of the wide variety of multimodal data it contains, which makes it a perfect fit for brain age estimation but as a next step we also plan to evaluate our framework on different tasks and datasets.

\subsubsection{Acknowledgements} KMB would like to acknowledge funding from the EPSRC Centre for Doctoral Training in Medical Imaging (EP/L015226/1).

% \newpage
\bibliographystyle{splncs04}
\bibliography{bibliography}

\end{document}

%% file: acronyms.tex
\DeclareAcronym{mlp}{
  short=MLP,
  long=Multi-Layer Perceptron
  }

\DeclareAcronym{gcn}{
  short=GCN,
  long=Graph Convolutional Network
  }

\DeclareAcronym{gnn}{
  short=GNN,
  long=Graph Neural Network
  }

\DeclareAcronym{ukbb}{
short=UKBB,
long=UK Biobank
}